\documentclass{article}

\usepackage{microtype}
\usepackage{graphicx}
\usepackage{subfigure}
\usepackage{booktabs} 

\usepackage{gensymb}
\usepackage{hyperref}

\usepackage[accepted]{icml2024}

\usepackage{amsmath}
\usepackage{amssymb}
\usepackage{mathtools}
\usepackage{amsthm}
\usepackage{multirow}

\usepackage{wrapfig}
\usepackage{caption}

\usepackage[capitalize,noabbrev]{cleveref}

\theoremstyle{plain}

\theoremstyle{definition}

\theoremstyle{remark}

\usepackage[textsize=tiny]{todonotes}
\usepackage{makecell}

\def\eg{\emph{e.g.}} 

\def\ie{\emph{i.e.}}

\usepackage{xspace}
\newcommand{\model}{Dual3D}
\newcommand{\modelname}{Dual3D\xspace}

\icmltitlerunning{\model: Efficient and Consistent Text-to-3D Generation with Dual-mode Multi-view Latent Diffusion}

\begin{document}

\twocolumn[
\icmltitle{\model: Efficient and Consistent Text-to-3D Generation with \\ Dual-mode Multi-view Latent Diffusion}




\icmlsetsymbol{inter}{\#}

\begin{icmlauthorlist}
\icmlauthor{Xinyang Li}{xmu}
\icmlauthor{Zhangyu Lai}{xmu}
\icmlauthor{Linning Xu}{cuhk}
\icmlauthor{Jianfei Guo}{pjlab}
\icmlauthor{Liujuan Cao}{xmu}
\icmlauthor{Shengchuan Zhang}{xmu}
\icmlauthor{Bo Dai}{pjlab}
\icmlauthor{Rongrong Ji}{xmu}
\end{icmlauthorlist}

\icmlaffiliation{xmu}{Key Laboratory of Multimedia Trusted Perception and Efficient Computing,
Ministry of Education of China, Xiamen University}
\icmlaffiliation{cuhk}{The Chinese University of Hong Kong}
\icmlaffiliation{pjlab}{Shanghai Artificial Intelligence Laboratory. Work done during an internship of Xinyang Li with Shanghai Artificial Intelligence Laboratory}

\icmlcorrespondingauthor{Liujuan Cao}{caoliujuan@xmu.edu.cn}

\icmlkeywords{Text-to-3D Generation; Latent Diffusion Models}

\vskip 0.3in
]



\printAffiliationsAndNotice{}  

\begin{abstract}
We present \modelname, a novel text-to-3D generation framework that generates high-quality 3D assets from texts in only $1$ minute.
The key component is a dual-mode multi-view latent diffusion model.
Given the noisy multi-view latents, the 2D mode can efficiently denoise them with a single latent denoising network,
while
the 3D mode can generate a tri-plane neural surface for consistent rendering-based denoising.
Most modules for both modes are tuned from a pre-trained text-to-image latent diffusion model to circumvent the expensive cost of training from scratch.
To overcome the high rendering cost during inference, we propose the dual-mode toggling inference strategy to use only $1/10$ denoising steps with 3D mode, successfully generating a 3D asset in just $10$ seconds without sacrificing quality.
The texture of the 3D asset can be further enhanced by our efficient texture refinement process in a short time.
Extensive experiments demonstrate that our method delivers state-of-the-art performance while significantly reducing generation time.
%
Our project page is available at \url{https://dual3d.github.io}.
\end{abstract}

\begin{figure*}[!t]
  \centering
  \includegraphics[width=1\linewidth]{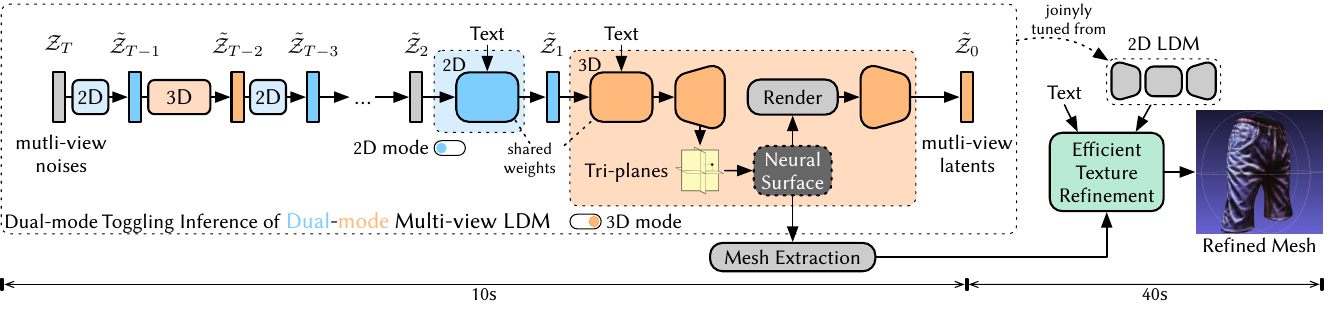}
  \caption{
  The Framework of \model.
  Firstly, we fine-tune a pre-trained 2D LDM into a dual-mode multi-view LDM.
  Subsequently, we employ a dual-mode toggling inference strategy to choose different denoising modes during inference to balance the inference speed and 3D consistency.
  Finally, the mesh extracted from the neural surface is further optimized via our efficient texture refinement process, enhancing the photo-realism and details of the asset.
  }
  \label{fig.1}
  \end{figure*}

\section{Introduction}


3D generation is a significant topic in the computer vision and graphics fields, which boasts wide-ranging applications across diverse industries, including gaming, robotics, and VR/AR.
%
With the rapid development of the 2D diffusion models, DreamFusion~\cite{poole2022dreamfusion} introduces Score Distillation Sampling (SDS) to use a pre-trained text-conditioned 2D diffusion model~\cite{saharia2022photorealistic} for generating 3D assets from open-world texts.
%
However, owing to the absence of 3D priors in 2D diffusion models, these methods frequently encounter low success rates and multi-faceted Janus problem~\cite{poole2022dreamfusion}.
On a different trajectory, direct 3D diffusion models~\cite{tang2023volumediffusion} offer alternative text-to-3D approaches with the denoising of 3D representations, but they always struggle with incomplete geometry and blurry textures due to the quality disparity between images and 3D representations.

To solve the multi-faceted Janus problem and generate high-quality assets with 3D-consistency, multi-view diffusion has garnered increasing interest since it can incorporate the rich knowledge of multi-view datasets.
Representative methods, MVDream~\cite{shi2023mvdream} and DMV3D~\cite{xu2023dmv3d}, introduce multi-view supervision into 2D and rendering-based diffusion models, respectively.
Specifically, MVDream fine-tunes a 2D latent diffusion model (LDM)~\cite{rombach2022high} into a multi-view latent diffusion model using multi-view image data, enabling efficient denoising of multi-view images. However, it still necessitates a time-consuming per-asset SDS-based optimization process to generate a specific 3D asset.
Conversely, DMV3D leverages multi-view diffusion in combination with a Large Reconstruction Model (LRM)~\cite{hong2023lrm}, enabling to generate a clean 3D representation during denoising without additional per-asset optimization.
Nevertheless, its denoising speed is inferior to MVDream as it necessitates full-resolution rendering at each denoising step.
Moreover, DMV3D trains the entire LRM from scratch, leading to a substantial increase in training cost relative to MVDream.

\begin{figure*}[!t]
  \centering
  \includegraphics[width=1\linewidth]{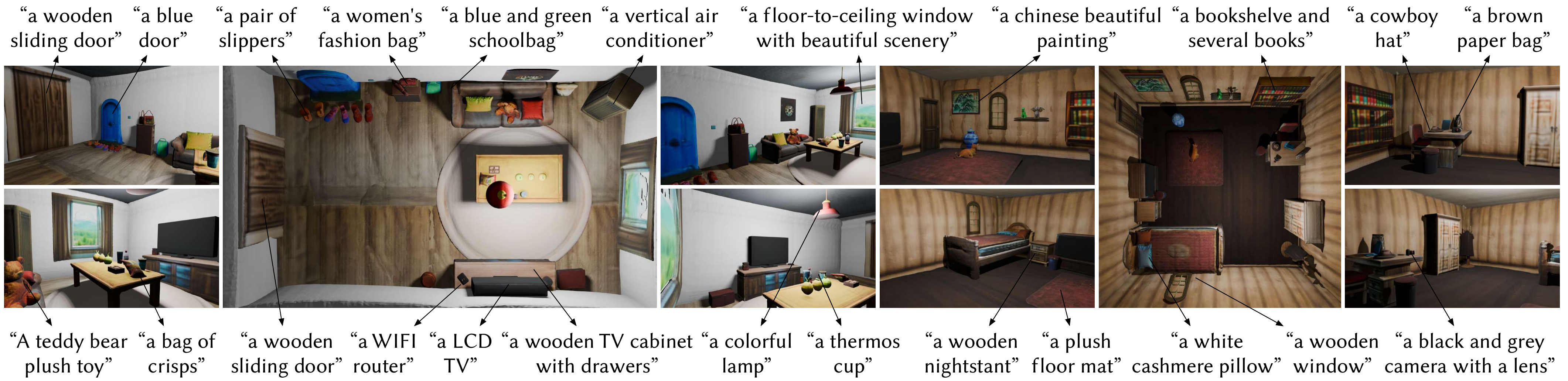}
  \caption{
  Two compositional 3D scenes rendered by Blender, where all visible assets are generated by our method with only texts as inputs.
  The text prompts for some assets are indicated by arrows.
  Please refer to our project page for the tour videos.
  }
  \label{fig.0}
  \end{figure*}

Our goal is to develop a high-quality text-to-3D generation framework with fast generation speed and reasonable training cost.
The cornerstone of our framework is a dual-mode multi-view latent diffusion model.
Both modes are trained with shared modules and only multi-view images as supervision.
During inference, we can toggle to 2D mode to reduce the inference time, or to 3D mode to obtain a noisy-free 3D neural surface for 3D-consistent multi-view rendering.
%
Also, to avoid the high training cost, we leverage the unified formulation of 2D latent features and 3D tri-plane features to design a novel architecture and training process that allows the dual-mode multi-view LDM to be tuned from a pre-trained 2D LDM.
The insight of this architecture is to replace the single-view latent denoising with synchronized and interconnected denoising of the multi-view latents and tri-plane representations.
We further discover that the texture of the 3D assets generated by only denoising exhibits a noticeable style difference from real-world textures, primarily due to the style bias in the synthesized multi-view datasets. To address this, we propose an efficient texture refinement process that quickly optimizes the texture map of the extracted mesh from the 3D neural surface.
The overall framework of our method is shown in Figure~\ref{fig.1}
and the entire generation process of our framework requires only $50$ seconds per asset on a single NVIDIA RTX 3090 GPU.
The efficient and high-quality generation ability makes our framework well-suited for compositional generation, and all generated 3D assets can seamlessly integrate into traditional rendering engines, as shown in Figure~\ref{fig.0}.

\section{Related Works}

\noindent \textbf{Text-to-3D Generation.}
DreamField~\cite{jain2022zero} pioneers the open-world text-to-3D generation domain by integrating vision language model CLIP~\cite{radford2021learning} with NeRF-based~\cite{mildenhall2021nerf} 3D rendering. 
DreamFusion~\cite{poole2022dreamfusion} and SJC~\cite{wang2023score} introduce 2D image diffusion models to optimize the 3D representation with SDS loss, improving the visual quality of text-to-3D generation.
With advancements in 3D representation~\cite{liu2020neural,chen2022tensorf} and rendering techniques~\cite{wang2021neus,yariv2021volume,tang2023dreamgaussian}, there has been a growing focus on extending these techniques to the text-to-3D generation domain. Notably, recent works~\cite{lin2023magic3d,tang2023dreamgaussian,liu2023unidream} have specifically targeted this area.
Furthermore, some methods propose alternative score distillation losses ~\cite{wang2023prolificdreamer,katzir2023noise,zou2023sparse3d,bahmani20234d,wu2024hd} to better leverage 2D diffusion models for stabilizing text-to-3D generation.
There are also methods~\cite{shi2023mvdream,liu2023unidream,long2023wonder3d,liu2023syncdreamer} that propose to introduce additional 3D priors to 2D diffusion models to improve the stability and 3D consistency of the generation.
However, SDS-based methods often require a long optimization time for each asset, making it challenging to apply to large-scale generation.

On the other hand, some approaches accomplish this task by directly training 3D generative models.
Early works~\cite{chan2021pi,schwarz2020graf,gu2021stylenerf,or2022stylesdf,chan2022efficient,deng2022gram,xiang2023gram} combine neural rendering and GANs~\cite{goodfellow2020generative,karras2018progressive,karras2019style,karras2020analyzing} techniques, yet their applicability is limited to specific categories.
High-capacity diffusion model methods~\cite{nichol2022pointe,jun2023shape,tang2023volumediffusion,shue20233d,gupta20233dgen,xu2023dmv3d}, nevertheless, either rely on 3D datasets or necessitate the reconstruction of multi-view datasets into 3D representations, resulting in high pre-processing cost.
These methods often encounter geometric artifacts and unrealistic textures due to the inherent discrepancy between 3D datasets and real-world images.

Our framework, \modelname, aims to generate high-quality and realistic 3D assets for category-agnostic texts while reducing the generation time to less than 1 minute.

\section{Preliminary}

\noindent \textbf{Latent diffusion models (LDMs)}~\cite{rombach2022high,saharia2022photorealistic}
consist of two key components: an auto-encoder~\cite{kingma2022autoencoding} and a latent denoising network.
The autoencoder establishes a bi-directional mapping from the space of the original data to a low-resolution latent space:
\begin{equation}
z = E(x), x = D(z),
\label{eq.1}
\end{equation}
where $E$ and $D$ are the encoder and decoder, respectively. 
The latent denoising network $\Tilde{\epsilon}_\theta$ is designed to denoise noisy latent given a specific timestep $t$ and condition $y$. 
Its training objective for $\epsilon$-prediction is defined as:
\begin{equation}
L=\mathbb{E}_{
x,\epsilon\sim\mathcal{N}(0,1),t}
\Big[\|\epsilon-\Tilde{\epsilon}_\theta(z_t,y,t)\|_2^2\Big],
\label{eq.2}
\end{equation}
where the noisy latent is obtained by $z_t = \sqrt{\bar{\alpha}_t} E(x) + \sqrt{1 - \bar{\alpha}_t} \epsilon$ and $\bar{\alpha}_t$ is a monotonically decreasing noise schedule.
During inference, a random noise is sampled as $z_T \sim \mathcal{N}(0,1)$. 
By continuously denoising the random noise $z_T$ with condition $y$, we can derive a fully denoised latent $\tilde{z}$.
Then, the denoised latent $\tilde{z}$ is fed into the latent decoder $D$ to generate the high-resolution image $\tilde{x} = E(\tilde{z})$.

\noindent \textbf{Multi-view diffusion models} aim to model the distribution of multi-view images $\mathcal{X}$ with 3D consistency, where each image is captured by a different camera within the same scene.
Its objective for $x_0$-prediction can be written as:
\begin{equation}
L=\mathbb{E}_{
\mathcal{X},\epsilon\sim\mathcal{N}(0,1),t}
\Big[\|\mathcal{X} - 
\Tilde{\mathcal{X}}_\theta(\mathcal{X}_t,c,y,t))\|_2^2\Big],
\label{eq.2}
\end{equation}
where $c$ represents the camera parameters for the different views.
Early works~\cite{shi2023mvdream, liu2023unidream} in this field are based on 2D LDMs. 
They fine-tune the 2D LDMs by integrating cross-view connections between the multi-view images into the original single-view 2D LDMs, using multi-view data rendered from 3D datasets. 
These methods lack strict 3D consistency since there is no actual 3D representation during multi-view denoising.
They also require the use of SDS-based optimization in conjunction with the fine-tuned multi-view LDMs to generate 3D assets.
A more advanced approach, DMV3D~\cite{xu2023dmv3d}, employs a 3D reconstruction model to generate noise-free 3D representations and predict multi-view images from noisy multi-view inputs by a 3D-consistent rendering process.
This allows for 3D generation tasks to be achieved without any per-asset optimization during inference.
%
However, the efficiency of the rendering-based multi-view diffusion models is significantly reduced due to the necessity of performing rigorous rendering on full-resolution images.

\begin{figure*}[!t]
  \centering
  \includegraphics[width=1\linewidth]{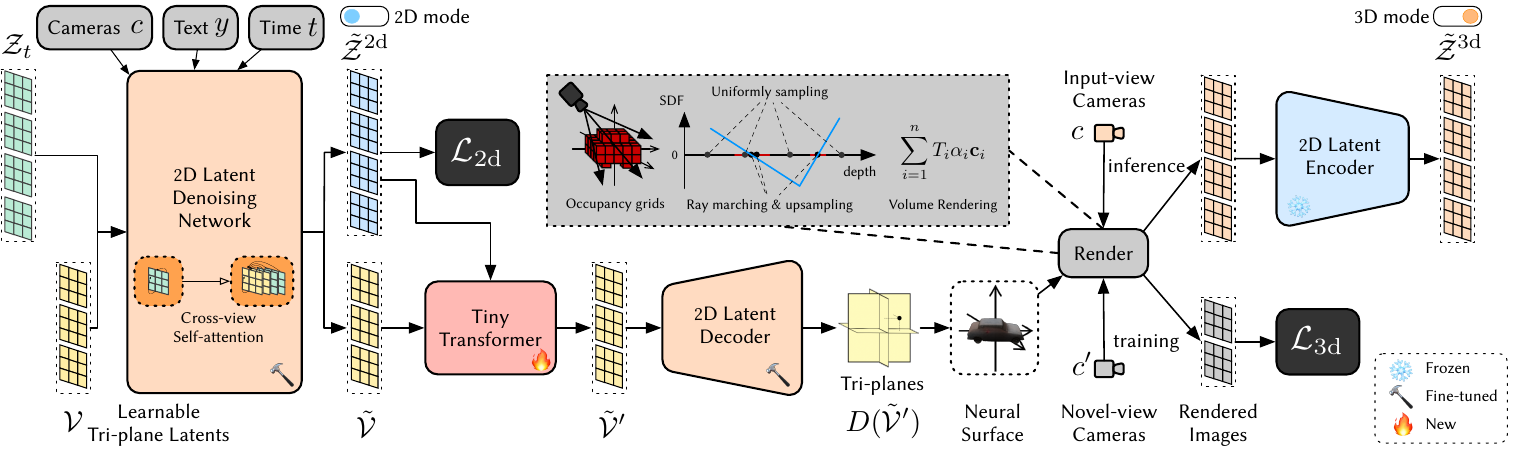}
  \caption{
  The architecture of dual-mode multi-view LDM. 
  The noisy multi-view latents and three learnable tri-plane latents are fed into the 2D latent denoising network $Z_\theta$ in parallel, where all self-attention blocks are replaced by cross-view self-attention blocks.
  A tiny transformer is used to enhance the connections between the multi-view features and the tri-plane features.
  The denoised tri-plane latents are decoded into higher resolution with the 2D latent decoder $D$ and rendered to images with volume rendering of the tri-plane surface.
  Two main objectives, $\mathcal{L}_{\text{2d}}$ and $\mathcal{L}_{\text{3d}}$, are used to optimize the model.
  }
  \label{fig.2}
\end{figure*}


\section{Method}

In this section, we outline the algorithm for tuning the pre-trained 2D LDM into the dual-mode multi-view LDM in Section~\ref{sec.4.1}.
We then introduce the dual-mode toggling inference strategy in Section~\ref{sec.4.2}.
Finally, we introduce the efficient texture refinement process in Section~\ref{sec.4.3}.



\subsection{Dual-mode Multi-view Latent Diffusion Model}
\label{sec.4.1}

The key insight of our approach is to utilize the strong prior of 2D LDM, (\ie, the compression and detail recovery abilities of the auto-encoder, and the generative ability of the latent denoising network) to jointly train a dual-mode multi-view LDM, where the 3D mode can directly generate a clean 3D representation from noisy multi-view latents.
We take multi-view images $\mathcal{X} \sim \mathbb{R}^{N\times 3\times H\times W}$ and feed them into the frozen image encoder $E$ of the 2D LDM to obtain the latents $\mathcal{Z} \sim \mathbb{R}^{N\times c\times h\times w}$.
During training, we add noise to the multi-view latents $\mathcal{Z}$ to derive the noisy latents $\mathcal{Z}_t = \sqrt{\bar{\alpha}_t} \mathcal{Z} + \sqrt{1 - \bar{\alpha}_t} \epsilon$, where $\epsilon$ is noise sampled from the Gaussian distribution $\mathcal{N}(0, 1)$ and $t$ is a random timestep.
One of our inspirations is that one of the popular 3D representations, tri-plane~\cite{chan2022efficient}, has a very similar formulation with 2D image features.
As such, we treat the tri-plane as three special latents.
Since we do not have the ground truth of the tri-plane latents, we initialize three learnable latents $\mathcal{V} \sim \mathbb{R}^{3 \times c \times h \times w}$ to serve as the noisy latents of the tri-planes, as illustrated in Figure~\ref{fig.2}.

\noindent \textbf{Denoising.}
Similar to the noisy multi-view latents $\mathcal{Z}_t$, the noisy tri-plane latents are also fed into the latent denoising network $Z_\theta$ in parallel, note that we change the $\epsilon$-prediction of the original 2D LDM into $x_0$-prediction for convenience.
The denoised tri-plane latents and multi-view latents can be obtained by $\Tilde{\mathcal{Z}}^{\text{2d}}, \Tilde{\mathcal{V}} = Z_\theta(\{\mathcal{Z}_t, \mathcal{V}\},c,y,t)$.
The camera condition $c$ is injected into the network by concentrating the rays $\boldsymbol{r}=(\boldsymbol{o}\times\boldsymbol{d},\boldsymbol{d})$,  parameterized by Plucker coordinates, into the noisy latents following LFN~\cite{sitzmann2021light}.
Here, $\boldsymbol{o}$ and $\boldsymbol{d}$ represent the origin and direction of the down-sampled pixel rays aligned with the latent resolution, respectively.
To make connections between the tri-plane features and multi-view features, we follow MVDream~\cite{shi2023mvdream} to simply replace self-attention blocks in the original 2D latent denoising network with cross-view self-attention blocks.
A straightforward multi-view latent diffusion objective is used to supervise the 2D-mode denoised multi-view latents $\Tilde{\mathcal{Z}}^{\text{2d}}$, which can be defined as:
\begin{equation}
\mathcal{L}_{\text{2d}}=\mathbb{E}_{
\mathcal{X},\epsilon,c,y,t}
\Big[\|\mathcal{Z} - \Tilde{\mathcal{Z}}^{\text{2d}}\|_2^2\Big].
\label{eq.2}
\end{equation}
We add a tiny transformer to further enhance the connections between the tri-plane latents and multi-view latents and get the final denoised tri-plane latents $\Tilde{\mathcal{V}}'$.
%
The final denoised tri-plane latents $\Tilde{\mathcal{V}}'$ are then fed into the latent decoder $D$ to get a high-resolution tri-planes $D(\Tilde{\mathcal{V}}')$, note that we re-initialize the last convolutional layer of the latent decoder $D$ for a higher number of tri-plane channels.

\noindent \textbf{Rendering.} Instead of using NeRF-based rendering, we follow TextMesh~\cite{tsalicoglou2023textmesh} to use NeuS~\cite{wang2021neus} as our base rendering method.
This allows for better geometric quality, and we propose some improvements for more efficient and accurate rendering.
First, we uniformly sample a certain resolution of dense grids within the bounding box and obtain the SDF values through bi-linear sampling of the tri-planes $D(\Tilde{\mathcal{V}}')$ and a tiny 2-layer MLP, following EG3D~\cite{chan2022efficient}. 
Then, we determine whether there could be a surface within the grid based on the SDF values of the grid center, marking the positive grids as occupied.
Next, for each ray, we obtain the initial sampled points within the occupancy grid via ray marching. 
These initial sampled points are refined through an upsampling strategy similar to NeuS, making the final sampled points close to the zero-set of the SDF.
We also concatenate some uniformly sampled points within the bounding box to explore unoccupied areas, the final color of the ray is calculated by volume rendering $\sum_{i=1}^nT_i\alpha_i\mathbf{c}_i$, where $T_i=\prod_{j=1}^n(1-\alpha_j)$, $n$ is the number of sampled points along the ray (sorted by depth), and $\alpha_i$ and $\mathbf{c}_i$ are the transparency and color of point $\mathbf{p}_i$, respectively.
The transparency $\alpha_i$ is calculated by $\alpha_i=\max(\frac{\Phi_s(f(\mathbf{p}_i))-\Phi_s(f(\mathbf{p}_{i-1}))}{\Phi_s(f(\mathbf{p}_i))},0)$, where $f(\cdot)$ is the SDF value of a specific point and $\Phi_s$ is the Sigmoid function with a learnable inverse standard deviation.
During training, we use novel-view cameras $c'$ instead of input-view cameras $c$ to supervise 3D-mode denoising in image space by:
\begin{equation}
\mathcal{L}_{\text{3d}}=\mathbb{E}_{\mathcal{X}',
\mathcal{X},\epsilon\sim\mathcal{N}(0,1),t}
\Big[ \ell(\mathcal{X}', 
R(D(\Tilde{\mathcal{V}}'), c')) \Big],
\label{eq.2}
\end{equation}
where $R$ is the rendering process, $D(\Tilde{\mathcal{V}}')$ is the tri-planes, $\ell(\cdot,\cdot)$ is an image reconstruction loss penalizing the difference between images, and $\mathcal{X}'$ is the ground truth of the novel-view images.
We use a combination of MSE loss and LPIPS~\cite{zhang2018unreasonable} loss with equal weights for the reconstruction loss $\ell$.
During inference, the images are rendered with the input-view cameras $c$ and encoded by the latent encoder $E$ to obtain the 3D-mode denoised latents, represented as $\Tilde{\mathcal{Z}}^{\text{3d}} =E(R(D(\Tilde{\mathcal{V}}'), c))$.
To regularize the surface to be physically valid and reduce floating geometry, we follow StyleSDF~\cite{or2022stylesdf} to employ the eikonal loss $\mathcal{L}_{\text{eik}}=(\|\nabla f(\mathbf{p})\|_2-1)^2$ and the minimal surface loss $\mathcal{L}_{\text{surf}}=\exp(-64| f(\mathbf{p})|)$ as constraints on the normal vectors and SDF values of sampled points $\mathbf{p}$.

\noindent \textbf{Total loss.} 
Our total loss is the weighted sum of the above losses, which is:
\begin{equation}
\mathcal{L}=
\lambda_{\text{2d}}\mathcal{L}_{\text{2d}} +
\lambda_{\text{3d}}\mathcal{L}_{\text{3d}} +
\lambda_{\text{eik}}\mathcal{L}_{\text{eik}} +
\lambda_{\text{surf}}\mathcal{L}_{\text{surf}},
\label{eq.4}
\end{equation}
where $\lambda_{\text{2d}}$, $\lambda_{\text{3d}}$, $\lambda_{\text{eik}}$, and $\lambda_{\text{surf}}$ are the weights of different losses.
We empirically set them to be $1$, $1$, $0.1$, and $0.01$, respectively, for all experiments.

\subsection{Dual-mode Toggling Inference}
\label{sec.4.2}

Our model is capable of performing both 2D-mode and 3D-mode denoising for multi-view latent diffusion.
Since the inputs and outputs of the two modes are perfectly aligned, we can toggle between them during inference.
While 3D-mode denoising ensures strict 3D consistency, it is significantly slower than 2D-mode denoising as it requires rendering full-resolution images, similar to DMV3D.
However, using too few 3D-mode denoising steps can lead to 3D inconsistency in the multi-view latents, resulting in artifacts in the final 3D assets.
We also find that the 3D-mode denoising is more difficult to deal with unseen texts due to the limited multi-view dataset, while the 2D-mode denoising can better handle unseen texts and concept combinations, as it is closer to the original 2D LDM.
Therefore, we propose dual-mode toggling inference to balance the inference speed, generation quality, and 3D consistency.
Specifically, we toggle between 2D-mode and 3D-mode denoising at a certain frequency throughout the entire inference process:
\begin{equation}
  \Tilde{\mathcal{Z}} = \begin{cases}
    \Tilde{\mathcal{Z}}^{\text{3d}},& \text{if } (t - 1) \text{ mod } m = 0 \\
    \Tilde{\mathcal{Z}}^{\text{2d}},& \text{otherwise}
  \end{cases}
  \label{eq.4}
\end{equation}
where $t$ is the current timestep and $m \in \mathbb{N}^+$ is the frequency to use 3D mode.
The denoised multi-view latents $\Tilde{\mathcal{Z}}$ are then used to denoise $\Tilde{\mathcal{Z}_{t}}$ into less noisy latents $\Tilde{\mathcal{Z}}_{t-1}$ using the $x_0$-prediction formulation of diffusion.
This design also ensures that the final denoising step is 3D-mode.
%
We experimentally find that only $1/10$ of the denoising steps need to use 3D mode (\ie, $m =10$ when using $100$ steps with DDIM~\cite{song2022denoising}), which can essentially ensure 3D consistency and reduce the inference time to $10$ seconds.

\subsection{Efficient Texture Refinement}
\label{sec.4.3}

The final 3D-mode denoising step of our method often yields 3D assets with good geometric shapes, but due to the limitations of the synthesized multi-view dataset, the textures are not always realistic.
Therefore, we propose an efficient texture refinement process to further enhance the texture, while keeping the time cost reasonable.
Specifically, we first extract the original neural surface into a mesh model, fix its geometry, and convert its texture into a learnable texture map. 
Then, we use differentiable mesh rendering~\cite{Laine2020diffrast} to render the mesh into an image $\mathcal{I}$ with a random viewpoint.
The image is encoded into the latent space, perturbed with an annealing strength of noise at timestep $t$, and denoised using a multi-step denoising process $F(\cdot,y,t)$ with the original 2D latent diffusion model.
Finally, We optimize the texture map by constructing a reconstruction loss between the rendered image and the refined image decoded from the denoised latent, which is:
\begin{equation}
  \|\mathcal{I} - D(F( \sqrt{\bar{\alpha}_t} E(\mathcal{I}) + \sqrt{1 - \bar{\alpha}_t} \epsilon), y, t)\|_2^2.
  \label{eq.5}
\end{equation}
This process significantly enhances the texture quality in a short time, thanks to the good surface quality of the neural surface generated from denoising and the efficiency of differentiable mesh rendering.
This process is inspired by the second stage optimization of DreamGaussian~\cite{tang2023dreamgaussian}. 
Still, our method is more robust and concise as the geometry and texture generated by our denoising stage provide better initialization, avoiding the complex color back-projection process in DreamGaussian.

\section{Experiments}

\subsection{Settings}

\noindent \textbf{Implementation details.}
We train our dual-mode multi-view LDM using the Adam optimizer~\cite{KingBa15} with a constant learning rate of $5e^{-5}$ and $(\beta_1, \beta_2) = (0.9, 0.95)$. 
The resolutions of the images and latents are $256$ and $32$, respectively.
We use $4$ input views following the practice of DMV3D~\cite{xu2023dmv3d}.
During training, we render $4\times 128 \times 128$ image patches for supervision to save GPU memory.
The batch size is set to $128$. 
Training takes about $4$ days with $32$ NVIDIA Tesla A100 GPUs for $100$K iterations.
We use Stable Diffusion v2.1 as our initial model.
We use $1000$ steps during training with a cosine schedule and reduce it to $100$ steps with DDIM~\cite{song2022denoising} during inference.
The classifier-free guidance scale is $7.5$ for 2D-mode denoising.
For rendering, we use $24$ uniformly sampled and $24$ upsampled points for each ray with the implementation of batch-wise neural surface rendering in StreetSurf~\cite{guo2023streetsurf}.
We directly use rendered multi-view images provided by Zero123~\cite{liu2023zero} with Objaverse~\cite{deitke2023objaverse} dataset to train our model.
The text prompts are generated by Cap3D~\cite{luo2023scalable}.
Like MVDream, our model also supports regularization with 2D text-to-image datasets such as LAION~\cite{schuhmann2022laion} to enhance generalization.
For texture refinement, we use a learning rate of $1e^{-1}$ and a total of $100$ iterations.
The pre-defined timestep is annealing from $0.20T$ to $0.05T$, where $T$ is the total denoising steps.

\begin{table}[!t]
  \centering
  \caption{Quantitative comparison.}
  \resizebox{0.47\textwidth}{!}{
    \begin{tabular}{l|c|c|c|c}
      \toprule
      Method          & \makecell{CLIP\\Similarity~$\uparrow$}  & \makecell{CLIP\\R-Precision~$\uparrow$}& \makecell{Aesthetic\\Score~$\uparrow$} & \makecell{Generation\\Time~$\downarrow$}\\
      \midrule
      Point-E         &66.2&47.2&4.39&21s             \\
      Shap-E          &70.4&60.0&4.40&\textbf{8s}            \\
      VolumeDiffusion-I &59.6&18.6&4.03&12s             \\
      \midrule
      Ours-I  &\textbf{72.0}&\textbf{72.3}&\textbf{5.22}&10s             \\
      \midrule
      DreamGaussian   &65.1&31.9&5.09&3m\\
      MVDream         &69.8&56.7&5.27&45m \\
      VolumeDiffusion-II &63.0&32.4&4.17&8m             \\
      \midrule
      Ours-II  &\textbf{73.1}&\textbf{74.3}&\textbf{5.50}&\textbf{50s}          \\ 
      \bottomrule
      \end{tabular}
  }
  \label{tab.1}
\end{table}

\noindent \textbf{Baselines.}
We compare our method with state-of-the-art category-agnostic text-to-3D
generation approaches, including Point-E~\cite{nichol2022pointe}, Shap-E~\cite{jun2023shape}, DreamGaussian~\cite{tang2023dreamgaussian}, MVDream~\cite{shi2023mvdream}, and VolumeDiffusion~\cite{tang2023volumediffusion}.
Point-E and Shap-E are 3D diffusion models based on point clouds and implicit functions, respectively.
DreamGaussian and MVDream are optimization-based methods that utilize the prior of 2D and multi-view LDMs, respectively.
VolumeDiffusion is a recent work that employs a two-stage framework for text-to-3D generation, combining a 3D volume denoising stage and an SDS-based refinement stage.
For a fair comparison, all generated assets are converted into meshes and rendered with Blender for evaluation to avoid differences in quality caused by different rendering processes.
We categorize the compared methods into inference-only and optimization-based according to whether there is a per-asset optimization process during generation.
Therefore, we classify the denoising stage of VolumeDiffusion and our method as inference-only methods (\ie, VolumeDiffusion-I and Ours-I), and the refining stage as optimization-based methods (\ie, VolumeDiffusion-II and Ours-II), with independent evaluation.
We would also report the results of DMV3D once its code is released.


\subsection{Quantitative results.}
\noindent \textbf{CLIP score.}
We compute the CLIP Similarity and R-Precision~\cite{cliprprecision} to compare the alignment between assets generated by different methods and the texts. 
For each method, we use $36$ texts from Shap-E to generate 36 assets and render $24$ fixed views for each asset.
The CLIP Similarity is calculated by computing the average cosine distance between the CLIP~\cite{radford2021learning} embeddings of the rendered view and the text.
The R-Precision is calculated by computing the Top-1 Precision of the zero-shot classification of the $36$ groups.
The results are shown in Table~\ref{tab.1}.
Our method demonstrates superior semantic alignment capability compared to baseline models using only the denoising stage.
After texture refinement, our method achieves better performance on both metrics, demonstrating the texture-enhancing ability of the texture refinement.

\begin{figure}[!t]
  \centering
  \includegraphics[width=1\linewidth]{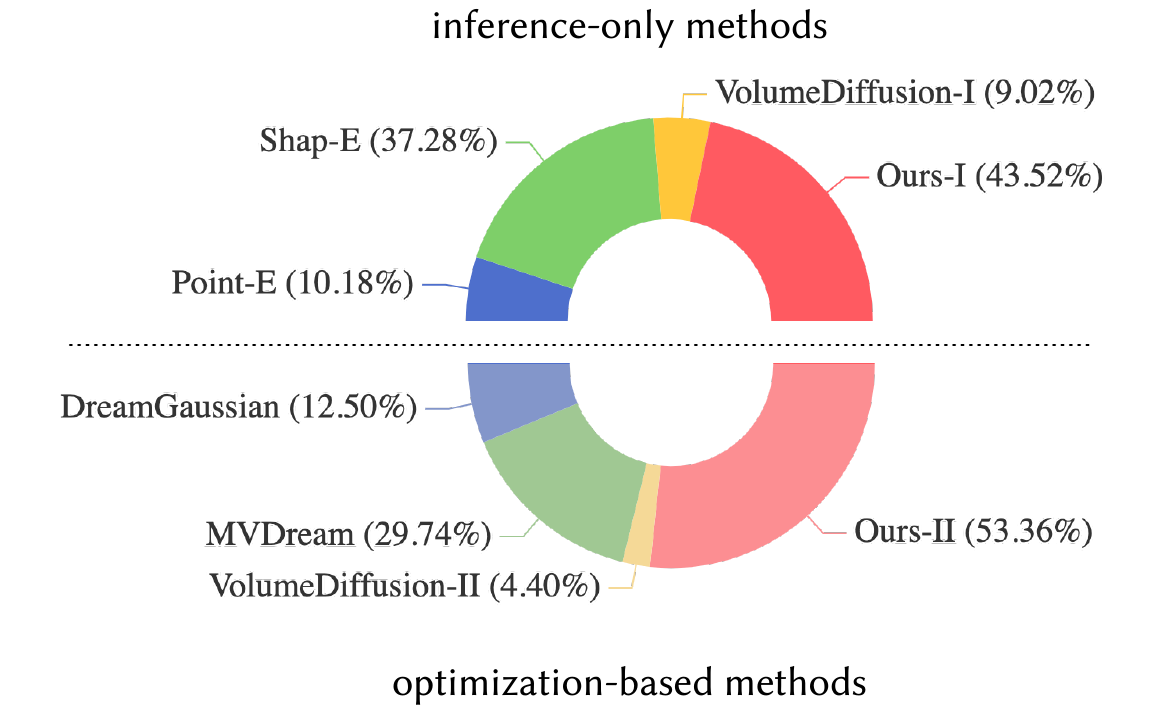}
  \caption{
  User study
  }
  \label{fig.4}
\end{figure}

\noindent \textbf{Aesthetic score.}
We also evaluate the aesthetic scores of the 3D assets generated by different methods.
We adopt the open-source LAION Aesthetics Predictor\footnote{https://github.com/LAION-AI/aesthetic-predictor}, which trains a single linear layer based on CLIP embeddings to predict the aesthetic quality of images from $0$ to $10$.
We use it to score the rendered images of the generated objects from the previous experiment and report the average.
The results are also shown in Table~\ref{tab.1}.
The results show that our method, using only the denoising stage, surpasses all other baseline methods except for MVDream.
After texture refinement, our method surpasses all baseline methods, demonstrating that our method can generate aesthetically pleasing 3D assets.

\begin{figure*}[!t]
  \centering
  \includegraphics[width=1\linewidth]{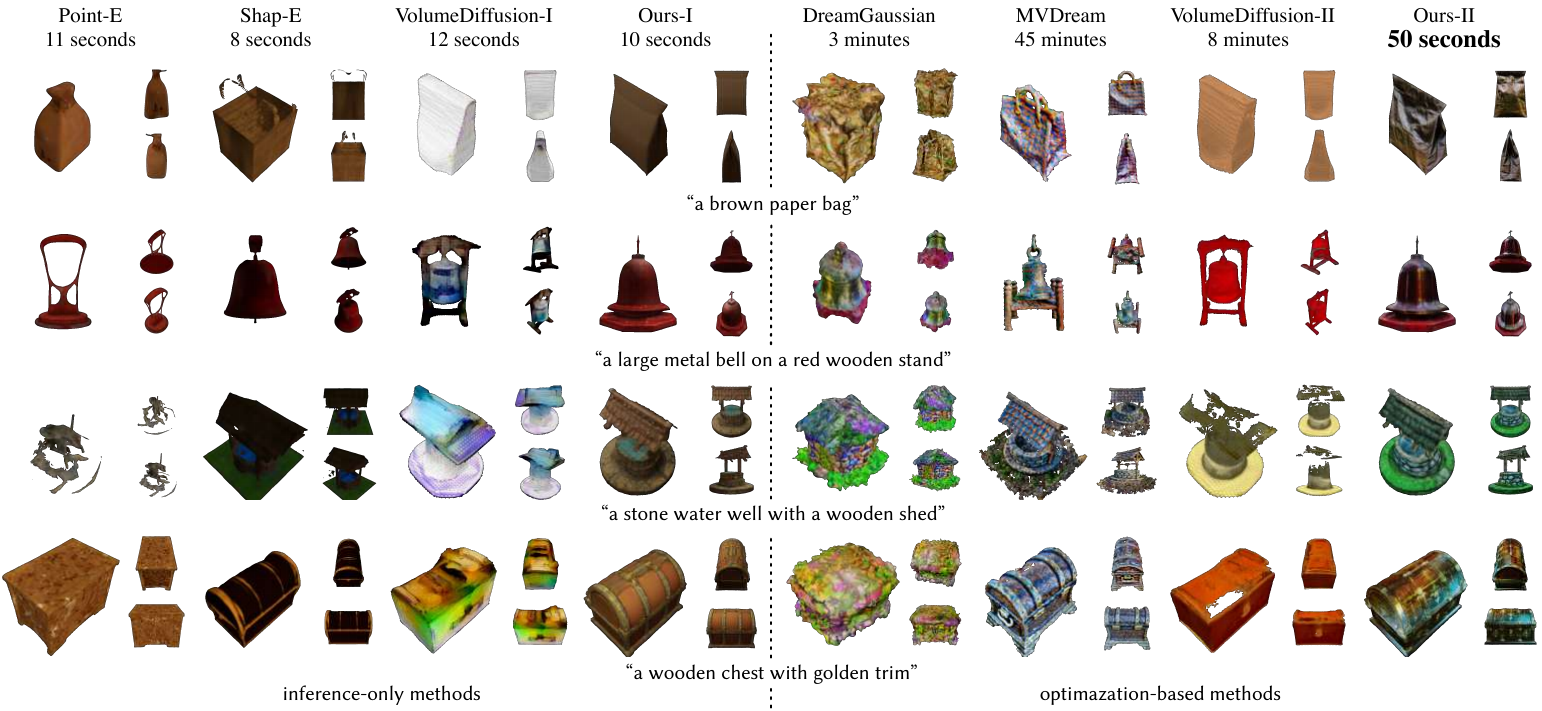}
  \caption{
  Qualitative comparison.
  }
  \label{fig.5}
  \end{figure*}

\begin{figure*}[!t]
  \centering
  \includegraphics[width=1\linewidth]{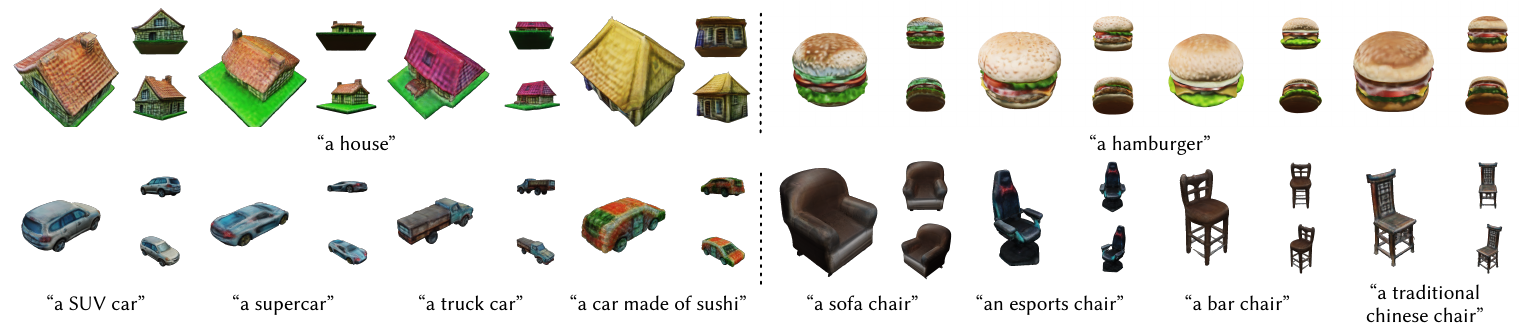}
  \caption{
  Diverse and fine-grained generation results.
  }
  \label{fig.6}
\end{figure*}

\noindent \textbf{Generation time.}
Considering application deployment and large-scale generation, we also evaluate the generation time of different methods.
For a fair comparison, we use a single NVIDIA RTX 3090 GPU to evaluate the generation time of different methods.
The results are reported in Table~\ref{tab.1}.
Point-E, Shap-E, VolumeDiffusion-I, and Ours-I are all inference-only methods, so the generation speed is fast.
DreamGaussian and MVDream are SDS-based methods, so the generation speed is slow.
Ours-II adopts an optimization-based refinement stage, but due to better initialization and the efficiency of mesh rendering, its speed is far superior to other optimization-based methods.

\noindent \textbf{User study.}
We also conduct a user study to compare the subjective quality of 3D assets generated by different methods.
We collect 36 votes from 24 users in 2 tracks (a total of 1728 votes) and count the percentage of votes obtained by each method, and the results are shown in Figure~\ref{fig.4}.
Different stages of our method all win the first place in the corresponding track, demonstrating that our method can generate 3D assets that align with user preferences.

\subsection{Qualitative results.}

We also qualitatively compare our method with baseline
methods as shown in Figure~\ref{fig.5}.
The inference-only and optimization-based methods are listed on the left and right of the dotted line, respectively.
For inference-only methods, Point-E, Shap-E, and VolumeDiffusion all produce discontinuous geometry, floating shapes, and poor texture.
Thanks to the image-space supervision and the effective utilization of the 2D prior model, our method can generate a complete shape and good texture that aligns with the text using only the denoising stage.
The good geometry and texture of the denoising stage also provide a better initialization for later refinement.
MVDream, as the method closest to our method in quantitative evaluation, also produces objects with realistic textures, but there are some holes in the geometry.
Although our method does not introduce explicit lighting or shadow during rendering, the strong prior of the 2D diffusion model helps to generate realistic lighting and shadow effects after texture refinement, making the generated 3D assets more photo-realistic.
Also, we find that the assets generated by our model are more consistent with the given text, especially in materials and colors, which aligns with our leading performance in the CLIP Score.

\subsection{Diverse and Fine-grained Generation}

In this experiment, we demonstrate some beneficial properties of our model.
On one hand, our model can generate diverse 3D assets given the same text. 
On the other hand, our model can generalize to fine-grained abstract semantic changes in the text. 
We select some texts and denoise different latent noises with different random seeds as shown in Figure~\ref{fig.6}. 
We first select some general sentences and generate four different objects. We find that our model can generate 3D assets that satisfy the given texts with varying contents, shapes, textures, and colors. 
We also select some base sentences and replace words in them.
Our model can translate these fine-grained semantic modifications into changes in the 3D geometry and shape details of the assets. 
This experiment demonstrates that our model has a strong ability for diverse and fine-grained generation.

\subsection{Ablation Study}

\noindent \textbf{\emph{w/o} network prior.}
In this ablation, we use a randomly initialized latent denoising network $\mathcal{Z}_\theta$ instead of resuming from the weights of the pre-trained 2D LDM.
Other settings remain the same as in the original model.
We present the quantitative metrics in Table~\ref{tab.2} to compare our full model with other methods.
We find that the model without prior experiences a significant drop in all metrics, especially for CLIP R-Precision. 
This demonstrates the effectiveness of training the dual-mode multi-view LDM with prior.

\begin{table}[!t]
  \centering
  \caption{Ablation study.}
  \resizebox{0.47\textwidth}{!}{
    \begin{tabular}{l|c|c|c|c}
      \toprule
      Method          & \makecell{CLIP\\Similarity~$\uparrow$}  & \makecell{CLIP\\R-Precision~$\uparrow$}& \makecell{Aesthetic\\Score~$\uparrow$} & \makecell{Generation\\Time~$\downarrow$}\\
      \midrule
      Ours-I  &72.0&72.3&5.22&10s             \\
      \midrule
      \emph{w/o} network prior &61.7&21.2&4.44&10s             \\
      \emph{w/o} tiny transformer &70.6&66.1&5.18&9s            \\
      \emph{w/o} dual-mode inference&66.0&44.6&4.81&1m30s            \\
      \bottomrule
      \end{tabular}
  }
  \label{tab.2}
\end{table}

\begin{figure}[!t]
  \centering
  \includegraphics[width=1\linewidth]{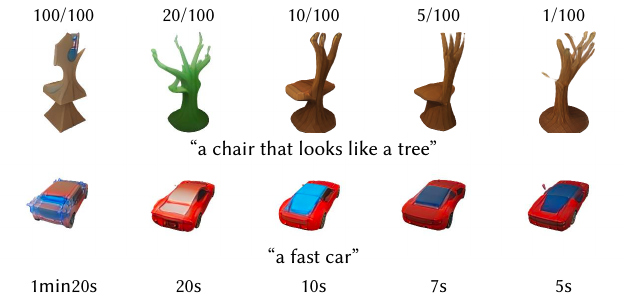}
  \caption{
  Ablation study on dual-mode toggling inference.
  }
  \label{fig.7}
\end{figure}

\noindent \textbf{\emph{w/o} tiny transformer.}
In this ablation, we remove the tiny transformer after the latent denoising network.
We find that this does not affect the overall convergence but has a significant adverse effect on the quality of the final 3D neural surface.
Because the original 2D latent denoising network uses a convolutional UNet~\cite{ronneberger2015u}, too little cross-view connection may make it difficult for the model to extract reasonable 3D information from the multi-view features.
As shown in Table~\ref{tab.2}, the CLIP R-Precision of this ablated model has a drop.
Also, since we use a dual-mode toggling inference, the overall generation time of this ablated model only has a slightly decreased.
Based on the above complaints, we think that introducing additional tiny transformers is necessary for our framework.

\noindent \textbf{\emph{w/o} dual-mode inference.}
In this ablation, we first remove the dual-mode toggling inference and use 3D mode for all denoising steps.
The metrics in Table~\ref{tab.2} show that the model performance has a significant decrease.
We then try different frequencies for 3D-mode denoising to observe the impact on efficiency and visual quality with a typical example shown in Figure~\ref{fig.7}.
The percentage of 3D-mode denoising and the inference time are on the top and the bottom, respectively.
We find that both too many and too few 3D-mode denoising steps lead to poor generation quality.
Too many 3D-mode steps lead to semantic misalignment, which we suspect is because the 3D-mode denoising is more difficult to utilize the original 2D prior, resulting in a poor understanding of complex semantics.
Too few 3D-mode steps lead to messy and floating geometry because the multi-view latents generated by 2D mode do not come from consistent 3D rendering.
Hence, we choose $10/100$ denoising steps to use 3D mode to basically ensure 3D consistency and generation quality with a reasonable time cost.

\begin{figure}[!t]
  \centering
  \includegraphics[width=1\linewidth]{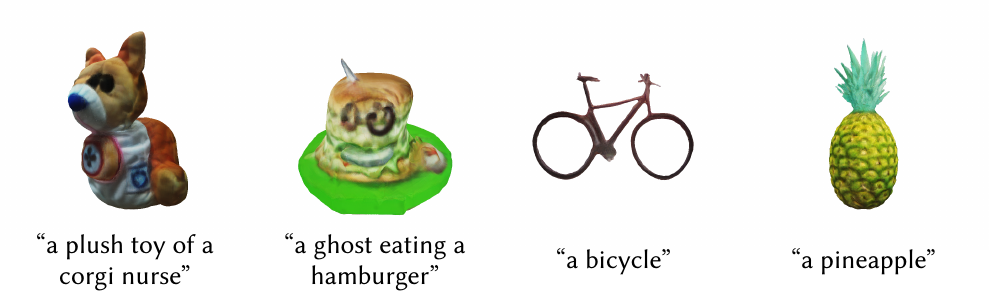}
  \caption{
  Some failure cases.
  }
  \label{fig.8}
  \vspace{-20pt}
\end{figure}

\subsection{Limitations}

Although our framework has a high success rate of generating in-distribution 3D assets and a certain generalization ability thanks to the prior of 2D LDM and the joint training of multi-view data and real-world 2D data, there are still some failure cases, as shown in Figure~\ref{fig.8}.
On the one hand, since most multi-view data are single-object scenes, it is difficult for our model to handle innovative text prompts with complex concepts or multi-object combinations (\eg, the ``eating'' action between the ghost and the hamburger is misunderstood).
This issue may be addressed by introducing more real-world multi-view data~\cite{yu2023mvimgnet,reizenstein2021common,ling2023dl3dv} and parameter-efficient fine-tuning techniques~\cite{hu2022lora,liu-etal-2022-p,he2022towards}.
On the other hand, although our texture refinement is efficient, using mesh rendering limits the further improvement of geometry, leading to the failure of generating very complex or thin shapes (\eg, The fine steel wire in the bicycle wheels and the small thorns on pineapple).
This issue may be addressed by introducing more efficient 3D representations, such as 3D Gaussian Splatting~\cite{kerbl3Dgaussians}, to further improve the rendering quality and efficiency.
These failure cases prompt us to further improve our framework design in our future works.




\section{Conclusion}

In this paper, we propose an efficient and consistent text-to-3D generation framework capable of generating realistic 3D assets in one minute.
We first introduce a dual-mode multi-view LDM that can be trained from a pre-trained 2D LDM, where the 3D mode can generate a clean neural surface from the noisy multi-view latents.
The proposed dual-mode toggling inference strategy further allows the dual-mode multi-view LDM to significantly improve the inference speed while ensuring consistency and generation quality.
The neural surface generated by the dual-mode multi-view LDM can be further extracted into a mesh and refined by the proposed efficient texture refinement to enhance the realism and details. 
We demonstrate the effectiveness of our method with extensive experiments and show the effect of each component.
We believe our work makes essential contributions to the text-to-3D generation community, especially in discovering the potential of dual-mode multi-view diffusion for fast and high-quality 3D generation.


%
%

\section*{Acknowledgements}
This work was supported by National Science and Technology Major Project (No. 2022ZD0118202), the National Science Fund for Distinguished Young Scholars (No.62025603), the National Natural Science Foundation of China (No. U21B2037, No. U22B2051, No. 62176222, No. 62176223, No. 62176226, No. 62072386, No. 62072387, No. 62072389, No. 62002305 and No. 62272401), and the Natural Science Foundation of Fujian Province of China (No.2021J01002,  No.2022J06001).

\bibliography{main}
\bibliographystyle{icml2024}

\newpage
\appendix
\onecolumn
\section{More Implementation Details.}

\subsection{Detailed Architecture of Dual-mode Multi-view LDM}

Here, we provide a more detailed explanation of our modifications in the 2D LDM, as shown in Table~\ref{tab.3}.
The latent encoder is completely frozen, so it does not require any modifications.
For the latent denoising network, we additionally input the camera condition in the form of Plucker coordinates, therefore increasing the channel dimension by $6$ in the input.
The first dimension is made up of the number of views and three triplanes.
We also increase the output of the latent denoising network and the input of the latent decoder by $508$ dimensions to match the difference in the amount of information between the tri-plane latents and the image latents.
The output of the latent decoder is increased from the original $3$ dimensions to $64$ dimensions, therefore the shape of the tri-planes is $3 \times 64 \times 256 \times 256$.

\begin{table}[!h]
  \centering
  \caption{Detailed Networks.}
  \resizebox{0.8\textwidth}{!}{
    \begin{tabular}{l|c|c|c}
      \toprule
      Module          & Architecture & Input& Output\\
      \midrule
      Latent Encoder  &CNN&$4\times 3\times 256 \times 256$&$4\times 4\times 32 \times 32$             \\
      Latent Denoising Network &UNet&$(4+3)\times (4+6)\times 32 \times 32$&$(4+3)\times (4+508)\times 32 \times 32$             \\
      Tiny Transformer &Transformer&$(4+3)\times (4+508)\times (32 \times 32)$&$3\times (4+508)\times (32 \times 32)$            \\
      Latent Decoder&CNN&$3\times (4+508)\times 32 \times 32$&$3\times 64\times 256 \times 256$            \\
      \bottomrule
      \end{tabular}
  }
  \label{tab.3}
\end{table}

\subsection{Tiny Transformer}

Our tiny transformer contains $16$ self-attention layers~\cite{vaswani2017attention}.
After each self-attention layer, there is a feed-forward MLP, which consists of two linear layers and a GeLU activation~\cite{hendrycks2016gelu}.
Layer Normalization~\cite{ba2016layer} is used before each self-attention and feed-forward layer.
We follow DIT~\cite{peebles2023scalable} to introduce zero-initialized layer scaling~\cite{goyal2017accurate} for each block to stabilize the training.
For attention layer, the numbers of the channels and heads are set to $512$ and $8$, respectively.
The parameter number is about $50$M.

\subsection{Adding Normalization Layers into Latent Decoder}

Our early experiment finds that although the loss of training the VAE decoder as the triplane upsampler is consistently decreased, it suddenly crashes during middle training.
By checking the data distribution of each layer output, we find that the main reason is that the output values of the decoder layers are gradually increasing, which leads to the numerical explosion.
We first try to add normalization layers (\eg, GroupNorm~\cite{wu2018group}) to each upsampling layer but find that it affects the original data distribution suddenly after initialization, which leads to slower convergence.
Finally, we adopt the exponential moving average normalization layer proposed by StyleGAN3~\cite{Karras2021} to each upsampling layer of the triplane decoder.
Specifically, we record the moving average norm $\sigma = \mathbb{E}(x^2)$ of each upsampling layer output $x$ and divide $x$ by $\sqrt{\sigma}$ for stabilizing.
$\sigma$ is initialized to be $1$, so it does not affect the distribution of each layer in the early stage, and it can be updated iteratively to stabilize the values during training.


\section{Evaluation Details.}

\subsection{CLIP Similarity and R-precision}

We directly use the text-to-3D evaluation from Cap3D~\cite{luo2023scalable} to calculate CLIP Similarity and R-precision with ViT-B/32 model.
For a fair comparison, we use $24$ fixed views (\ie, $\text{azimuth}\in \{0\degree,45\degree,90\degree,135\degree,180\degree,225\degree,270\degree,315\degree\}$ and $\text{elevation}\in \{-30\degree,0\degree,30\degree\}$) to render 3D assets from different methods.
The CLIP Similarity is calculated by
$
  \mathbb{E} [\cos{(\mathbf{f}_{x},\mathbf{f}_{y})} \times 100 \times 2.5]
$,
where $\mathbf{f}_{x}$ and $\mathbf{f}_{y}$ are the CLIP embeddings of the rendered image $x$ and text $y$, respectively.
For R-precision, we calculate the similarity of each rendered image and all $36$ texts.
Then, we report the proportion of all rendered images that the similarity with the ground truth text is the Top-1.

\subsection{User study}

\begin{figure}[!h]
  \centering
  \includegraphics[width=0.5\linewidth]{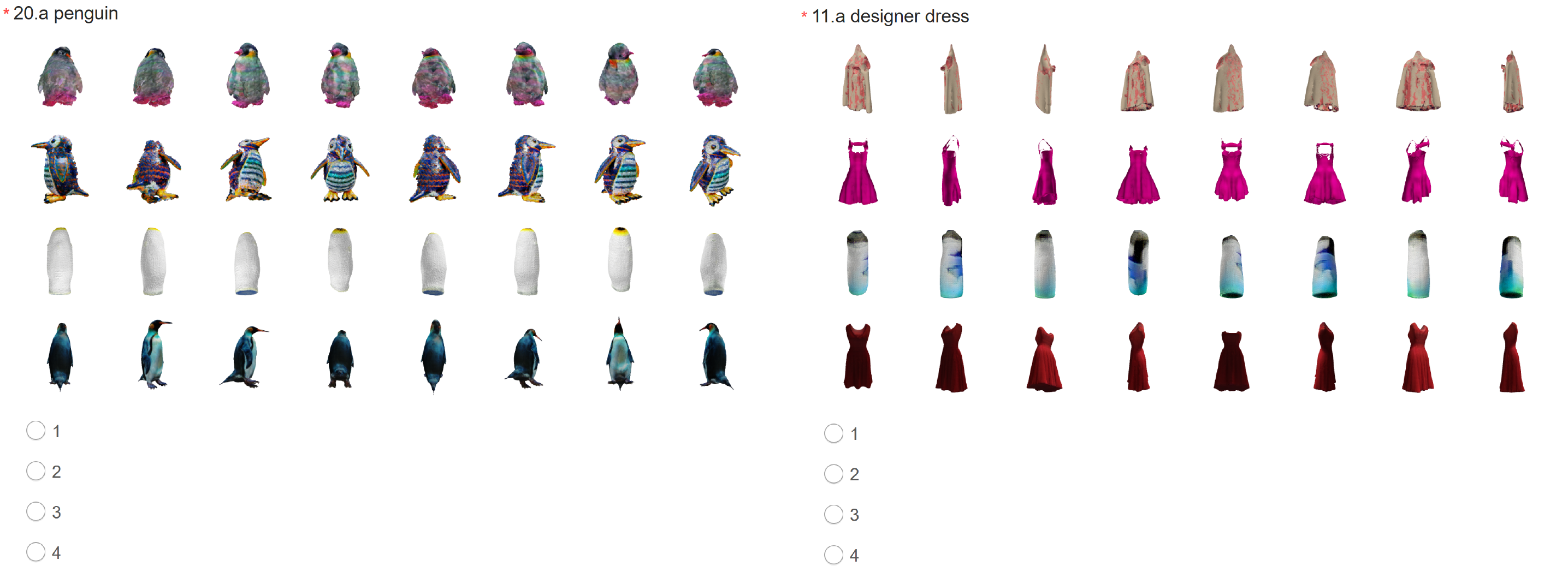}
  \caption{
  Examples of user study.
  }
  \label{fig.sup.0}
\end{figure}

The user study is conducted in the form of an anonymous questionnaire.
We invite $24$ users to participate, and each user must complete $36$ votes for two tracks.
We provide an example for both tracks, as shown in Figure~\ref{fig.sup.0}.
Each example requires users to choose the most satisfactory one from four different methods according to the text description.

\section{More Results.}

\subsection{Comparison with DMV3D}

\begin{figure}[!h]
  \centering
  \includegraphics[width=0.5\linewidth]{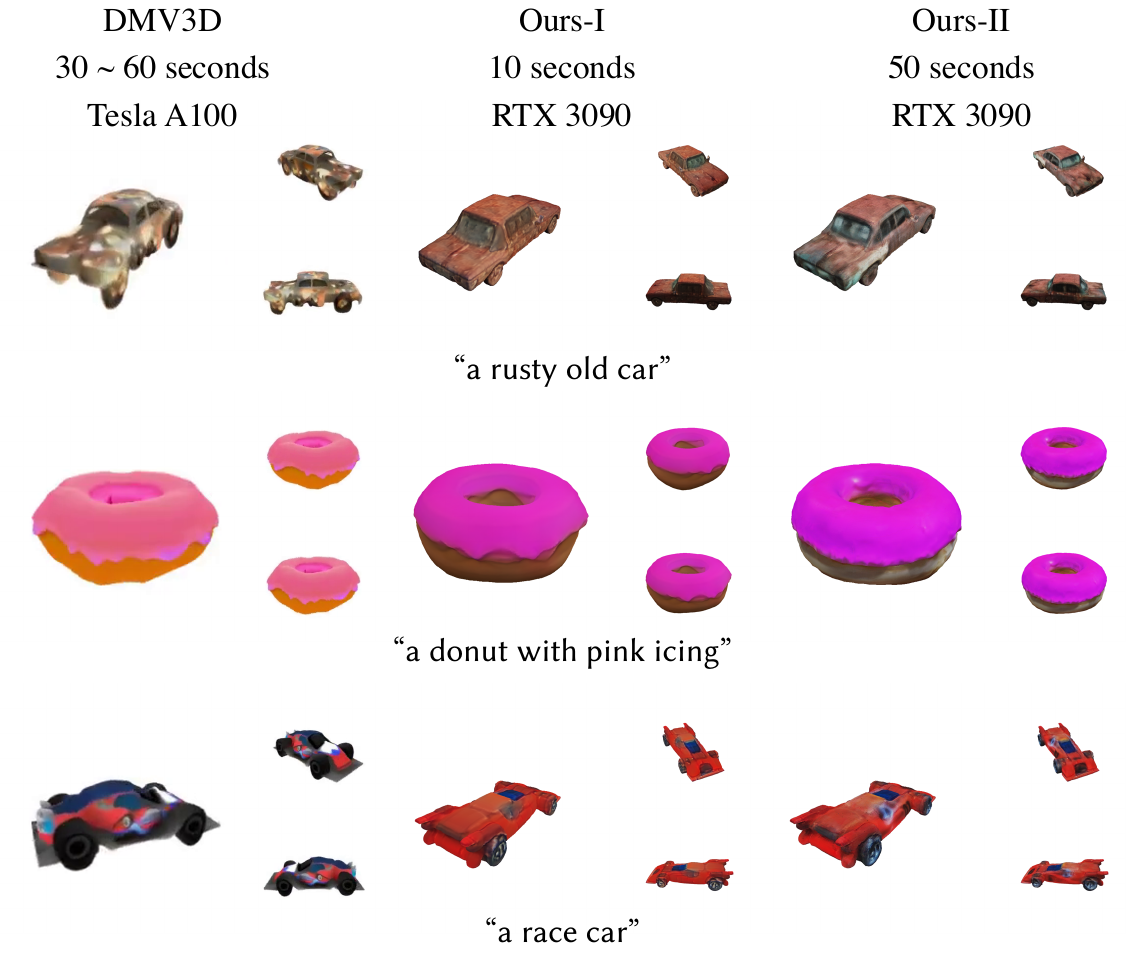}
  \caption{
  Qualitative comparison with DMV3D.
  }
  \label{fig.sup.3}
\end{figure}

Our framework has the following advantages compared to DMV3D:
1. We have greatly reduced the required training cost and time. 
DMV3D requires $128$ A100 cards to train for one week, while we only need $32$ cards to train for $4$ days, which is about $1/8$ GPU days of DMV3D. 
2. The proposed dual-mode toggling inference reduces the inference time of denoising. 
DMV3D takes approximately $30$ seconds to $1$ minute with an NVIDIA Tesla A100 GPU, while we only need $10$ seconds with an NVIDIA RTX 3090 GPU for denoising.
3. Our generated mesh after texture refinement is more realistic, and our model has more potential for generalization of unseen texts due to the effective utilization of the 2D LDM.
We supply some qualitative comparisons with DMV3D in Figure~\ref{fig.sup.3}.
Since the code of it has not been released, we directly use the examples on the project page.
We find that the textures generated by DMV3D are more abundant while our model tends to generate more details (\eg, the rearview mirrors of ``a rusty old car'' and the tires of ``a race car'').

\subsection{Additional Qualitative Comparison with Baselines}

\begin{figure}[!t]
  \centering
  \includegraphics[width=1\linewidth]{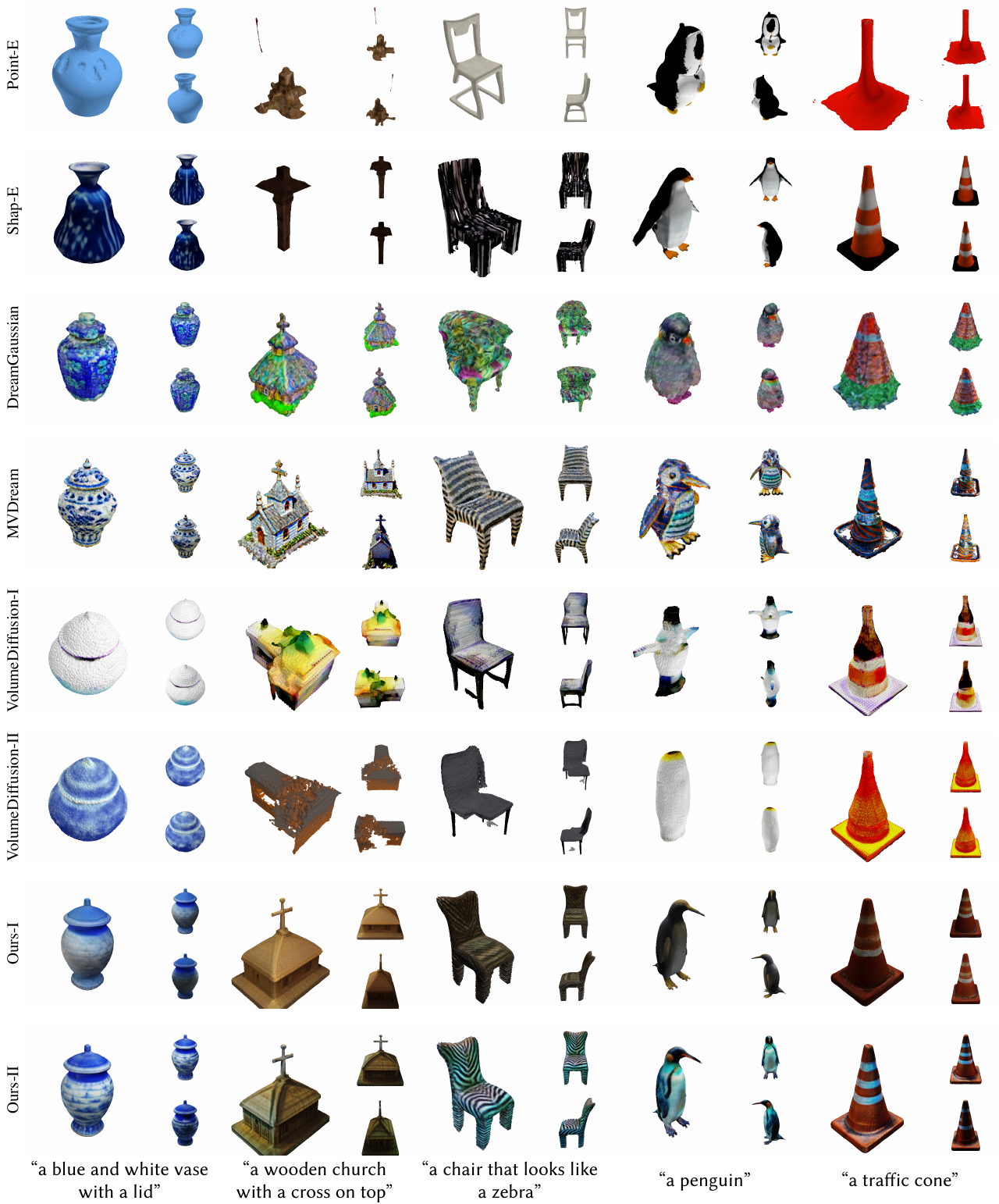}
  \caption{
  More qualitative comparison.
  }
  \label{fig.sup.1}
\end{figure}

We provide some additional qualitative comparisons with the baseline models in Figure~\ref{fig.sup.1}.
The text prompts are from Shap-E and VolumeDiffusion.

\subsection{Additional Visualization Results}

\begin{figure}[!t]
  \centering
  \includegraphics[width=1\linewidth]{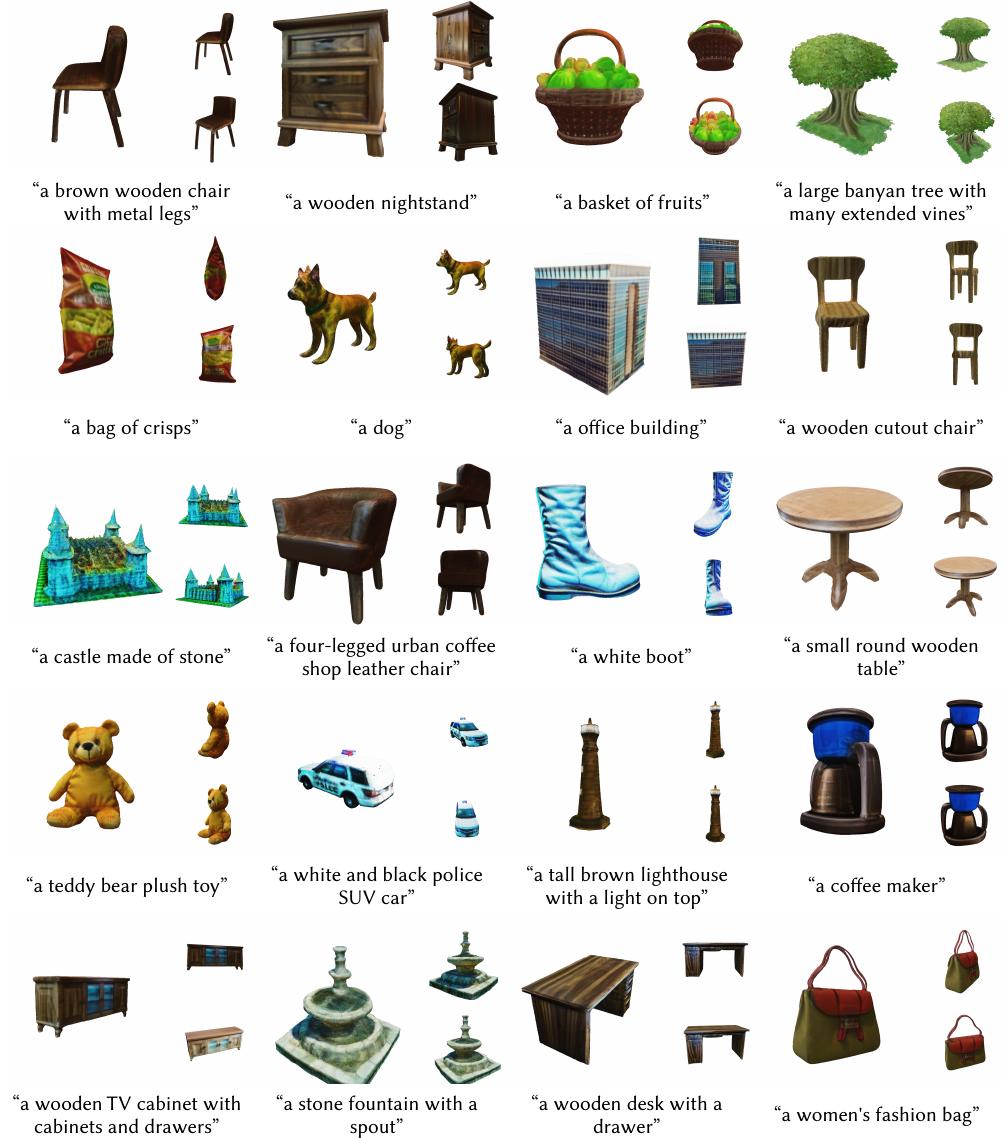}
  \caption{
  More results.
  }
  \label{fig.sup.2}
\end{figure}

We also provide some additional visualization results of our framework in Figure~\ref{fig.sup.2}.
For more interactive results, please kindly refer to our project page.

\end{document}